\documentclass[twoside]{article}
\usepackage[utf8]{inputenc}
\usepackage{subcaption}
\usepackage{amsmath}
\usepackage{amssymb}
\usepackage{hyperref}
\usepackage{titlesec}
\usepackage{xcolor}
\usepackage{fancyhdr}
\usepackage{graphicx}
\usepackage{multirow}
\usepackage[rightcaption]{sidecap}
\usepackage{verbatim}
\usepackage{listings}
\usepackage[backend=bibtex]{biblatex}
\usepackage [ a4paper , hmargin = 1 in , bottom =1 in ] { geometry }
\hypersetup{
    colorlinks=true,
    linkcolor=blue,
    filecolor=magenta,      
    urlcolor=blue,
    citecolor=black,
}

\begin{document}
\title{\textbf{Shayona@SMM4H’23: COVID-19 Self diagnosis classification using BERT and LightGBM models  }}
\author{Rushi Chavda$^1$, Darshan Makwana$^1$, Vraj Patel$^2$, Anupam Shukla$^2$  \\ $^1$Indian Institute of Technology, Bombay \\ $^2$National Institute of Technology, Surat }
\date{}
\maketitle
\vspace{1cm}

\section*{Abstract}
\begin{justify}
    \textit{This paper describes approaches and results for shared Task 1 and 4 of SMMH4-23 by Team Shayona. Shared Task-1 was binary classification of english tweets self-reporting a COVID-19 diagnosis, and Shared Task-4 was Binary classification of English Reddit posts self-reporting a social anxiety disorder diagnosis. Our team has achieved the highest f1-score 0.94 in Task-1 among all participants. We have leveraged the Transformer model (BERT) in combination with the LightGBM model for both tasks.  }
\end{justify}

\section*{Introduction}
\begin{justify}
    In recent years, the surge in worldwide social media utilization has provided researchers with fresh prospects to extract health-related data\cite{2}, which can ultimately be harnessed to enhance public health.
    The Social Media Mining for Health Applications (SMM4H) Shared Tasks address natural language processing (NLP) challenges of using social media data for health informatics, including informal, colloquial
    expressions, misspellings, noise, data sparsity, ambiguity, and multilingual posts\cite{8}. In its 2023 call, 5 shared tasks were given. We had participated in Task-1 and Task-4.
\end{justify}

\begin{justify}
    In Task-1, In the context of monitoring COVID-19 experiences via Twitter, this research addresses the challenge of automated binary classification. The task involves discerning self-reported COVID-19 diagnoses (”1”) from non-diagnostic mentions (”0”), enabling real-time, large-scale analysis. This problem focuses on distinguishing explicit diagnoses rather than mere user experiences.
\end{justify}

\begin{justify}
     Task-4 was also a binary classification problem, Exploring patients’ perspectives and medical needs
    through social media analysis offers insights into disease journeys. Social media listening (SML) has
    the potential to advance disease understanding and therapy development\cite{3}. For this task, we used a dataset
    extracted from the subreddit r/socialanxiety. The challenge is to build a classifier that correctly identifies
    patients that report having a positive or probable diagnosis of social anxiety disorder (positive cases
    labeled as ‘1’) from patients that report not having a diagnosis or the presence of a diagnosis is unlikely
    or unclear (negative cases labeled as ‘0’).
\end{justify}

\begin{justify}
    We know Transformer models are outperforming in all Natural Language downstream tasks\cite{7}, we have leveraged BERT model (digitalepidemiologylab/covid-twitter-bert-v2\cite{4}) from hugging face and further fine tuned it using given training data on binary classification task, and then used its embeddings to train LightGBM model. The selection of LightGBM is mostly a result of its enormous success since its launch in 2017. The benefits it has over other algorithms support this accomplishment\cite{6}. In fact, LightGBM is well-known for its rapid training speed, high accuracy, and minimal memory usage, as well as for supporting parallel, distributed, and GPU Learning\cite{6}. Going ahead in this paper we have shown comparison between our different approaches, and future work. Both tasks provide a binary classification problem, so we have used the same model architecture in both the tasks. In The Discussion section we have provided potential Improvements for Task-1 and Task-4. 
\end{justify}
\begin{figure}[!h]
    \centering
    \includegraphics[scale=0.5]{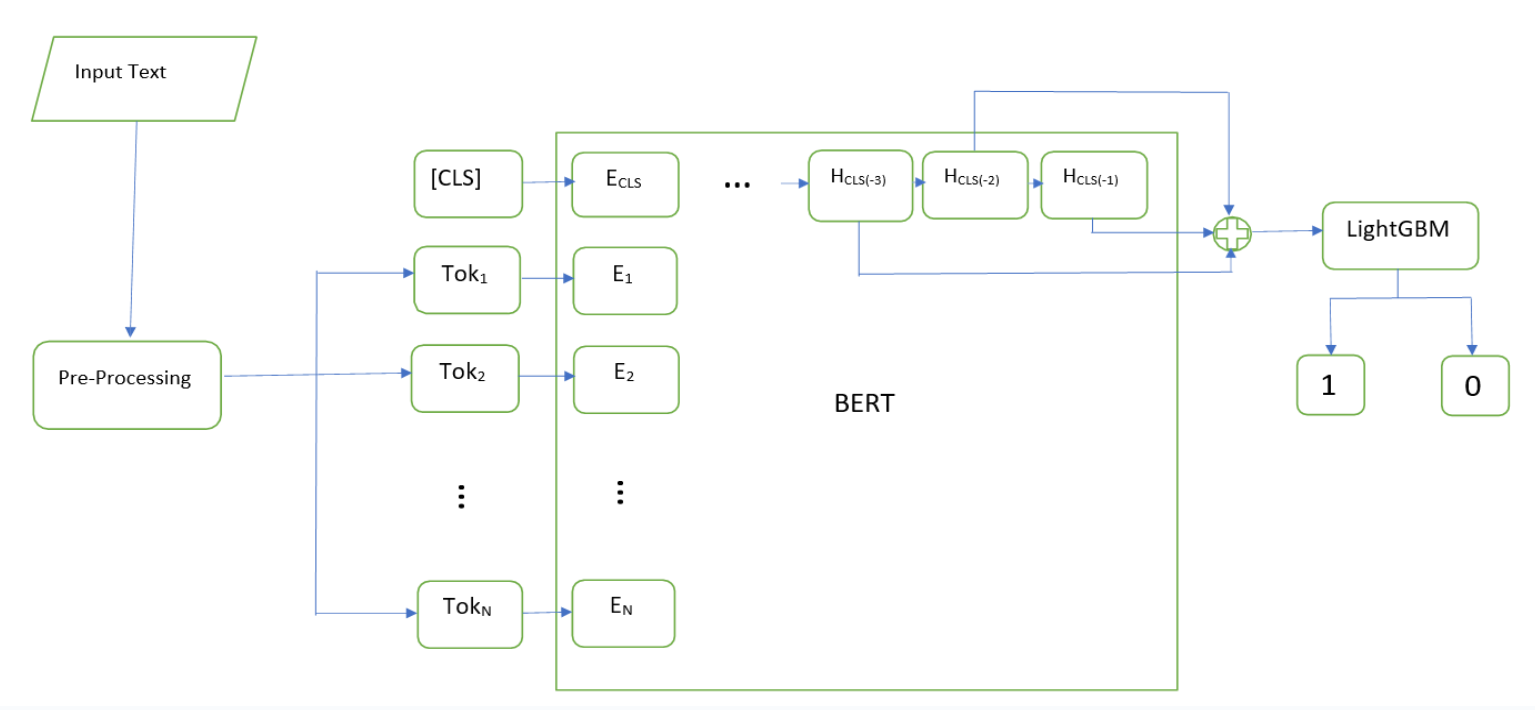}
    \caption{Figure 1: An overview of the proposed method}
    \label{fig1}
\end{figure}

\section*{Data}
\begin{justify}
In Task-1, our dataset comprised a total of 7,600 annotated tweets for training, 400 for validation, and 10,000 for testing our model. Each entry contained three columns: tweet ID, tweet text, and a binary label (0 or 1). Notably, within the training set of 7,600 tweets, 1,334 were labeled as 1, indicating the presence of self-reports of COVID diagnosis. 
\end{justify}

\begin{justify}
For Task-4, we worked with a dataset consisting of 6,090 annotated tweets for training, 680 for validation, and 1,347 for testing our model. Similar to Task-1, each entry featured three columns: ID, text, and a binary class label (0 or 1). In this dataset as well, 1,334 tweets were labeled as 1, denoting the occurrence of self-reports of COVID diagnosis. Data preprocessing steps were employed to clean and prepare the text data. These steps involved filtering out non-alphabet characters, tags, emojis, and URLs, as they typically carry minimal relevance for text comprehension. we also removed stopwords and punctuations. 
\end{justify}
\section*{System}
\begin{justify}
    We have used the BERT (bidirectional encoder representations from transformers) model from hugging face. We have used “digitalepidemiologylab/covid-twitter-bert-v2” BERT model from hugging face, which is specially trained on COVID-19 Twitter data and more suitable for Task-1\cite{5}. Our method leverages the power of BERT to capture complex linguistic patterns and the efficiency of LightGBM to optimize feature space and classification\cite{7}.
\end{justify}

\begin{justify}
    For Task-1, we conducted experiments with three methods
    \begin{enumerate}
        \item RoBERTa model for sequence classification\cite{8}
        \item BERT model fine-tuned for classification
        \item BERT model with the integration of LightGBM\cite{7}
    \end{enumerate}
    
    In the first method, we used pre-trained \textit{RobertaForSequenceClassification} model from hugging face\cite{9}, we used weights of roberta-base model, we fine-tuned the model with preprocessed training data, and using this, we achieved an f1-score of 0.83 on validation data.
\end{justify}
 
\begin{justify}
    In our second approach, we leveraged a pre-trained BERT model trained on COVID-related tweets\cite{5}. We extended this model with two fully connected layers: [1024, 256] and [256, 1]. These layers progressively distilled the features for our binary classification task. During training, we computed predictions through a forward pass and gauged their accuracy with BCE loss. In back propagation, gradients were calculated concerning the loss and propagated backward. Using the Adam optimizer, which merges adaptive learning rates and momentum, model parameters were adjusted according to their impact on loss reduction. This dynamic fine-tuning process enabled the model to refine its predictions iteratively. We achieved a f1-score of 0.89 on the validation set using this approach.
\end{justify}
 
\begin{justify}
LightGBM is a robust gradient-boosting framework that excels at handling complex tasks involving tree-based learning techniques. Combined with BERT, a model renowned for capturing intricate semantic information from text, we unlock the synergistic potential of both methods. BERT's strength lies in its ability to understand the meaning of words and phrases in context, while LightGBM's forte is navigating through high-dimensional data and optimizing complex problems.

In our third experimental approach, we seamlessly integrated LightGBM with the BERT model, as illustrated in Figure~\ref{fig1}. Here's how we did it step by step:

\begin{enumerate}
    \item \textbf{BERT Fine-Tuning:} First, we fine-tuned the BERT model following the same process as in our second experiment (Experiment 2). However, crucially, we did not freeze the model parameters during this phase. This allowed the BERT model's parameters to be updated and adapted to our specific task.
    
    \item \textbf{Hidden [CLS] Token Extraction:} Next, we extracted the Hidden [CLS] tokens from the last three layers of the BERT model. This choice was inspired by the approach outlined in the work of Eessa et al.~\cite{7}. These extracted tokens served as the input for our LightGBM model.
    
    \item \textbf{Hyperparameter Optimization:} We employed hyperparameter optimization techniques using the Optuna framework\cite{10} to ensure optimal performance. This step fine-tuned the parameters of our LightGBM model, enabling it to make the most accurate predictions.
\end{enumerate}

The results were impressive. Our LightGBM model and hyperparameter tuning outperformed all other methods we explored. Specifically, it achieved an F1 score of 0.93 on the validation dataset and an even higher F1 score of 0.94 on the test dataset for Task 1, making it the top-performing model for this task. Encouraged by this success, we extended our approach to Task 4, another binary classification problem. Here, our model delivered an F1-score of 0.8 on the test data, reaffirming its versatility and effectiveness.

\end{justify}

\section*{Results}

\begin{table}[!h]
\begin{center}
\begin{tabular}{|c|c|c|c|} 
 \hline
       & RoBERTa & BERT & BERT+LightGBM \\
 \hline
  Validation & 0.83 & 0.89 & \textbf{0.93} \\
 \hline
Test & \textbf{---} & \textbf{---} & \textbf{0.94} \\
 \hline
\end{tabular}
\caption{\label{tab1}Task-1 results for different experiments.}
\end{center}
\end{table}

\vspace*{-0.5cm}

\begin{justify}
    Table \ref{tab1} shows the performance of different models used in Task-1. With the hybrid model of BERT and LightGBM, we were able to reach a Precision, Recall, and f1-score of 0.949, 0.938, and 0.943 respectively on the test set.
\end{justify}
\begin{justify}    
    Since the hybrid model had the highest f1-score among all other methods in Task-1, we used the same for Task-4 as well, and since Task-1 and Task-4 both are binary classification problems, we kept the model architecture the same for Task-4 as well. This resulted in a Precision, Recall, and f1-score of 0.756, 0.871, and 0.809 respectively on the test set.
\end{justify}

\section*{Discussion}
\begin{justify}
    This paper presents the participation of the Shayona team in SMM4H’23. For both tasks the evaluative parameter of the competition was f1-score and We got highest f1-score in the Task-1. We observed that the gradient boosted algorithm (LightGBM) shows the best results. The LightGBM model can effectively aggregate the information from the concatenated [CLS] tokens. LightGBM is a gradient boosting framework that can learn complex interactions between features. It might be identifying patterns and relationships within the concatenated tokens that are not readily apparent from a single layer’s representation. 
\end{justify}

\begin{justify}
    Due to time and computation constraints, we were unable to explore more, but we could try  changing the number of layers from which we are extracting hidden [CLS] tokens. We could also try with different layer dimensions. Due to the fact that we took a hugging face BERT model which was pre-trained on a model that was trained using COVID-19-related tweeter data rather than Task-4 specific data, we were unable to attain a good f1-score in Task-4. If we employ a BERT model, pre-trained on Task-4 related data then the f1-score could be improved.
\end{justify}
\section*{Conclusions}
\begin{justify}
    Our research showcases a hybrid BERT and LightGBM model that excels in the binary classification problem of COVID-19 self-diagnosis (Task-1). By integrating BERT’s language understanding with LightGBM’s feature optimization, their approach achieved an impressive f1-score of 0.94 in Task-1, highlighting the synergy of transformer embeddings and gradient boosting for effective health-related classification. 
\end{justify}
\defbibheading{centered}{\section*{\centering References}}
\printbibliography[heading=centered]

@article{1,
  title={Social Media Use for Health Purposes: Systematic Review},
  author={Chen, J. and Wang, Y.},
  journal={Journal of Medical Internet Research},
  volume={23},
  number={5},
  pages={e17917},
  year={2021},
  doi={10.2196/17917},
}

@inproceedings{2,
  title={Overview of the eighth Social Media Mining for Health Applications (SMM4H) Shared Tasks at the AMIA 2023 Annual Symposium},
  author={Klein, A. Z. and Banda, J. M. and Guo, Y. and Flores Amaro, J. I. and Rodriguez-Esteban, R. and Sarker, A. and Schmidt, A. L. and Xu, D. and Gonzalez-Hernandez, G.},
  booktitle={Proceedings of the Eighth Social Media Mining for Health Applications (SMM4H) Workshop and Shared Task},
  year={2023},
}

@article{3,
  title={Evaluating Patient Experiences in Dry Eye Disease Through Social Media Listening Research},
  author={Cook, N. and Mullins, A. and Gautam, R. and Medi, S. and Prince, C. and Tyagi, N. and Kommineni, J.},
  journal={Ophthalmology and Therapy},
  volume={8},
  number={3},
  pages={407--420},
  year={2019},
  doi={10.1007/s40123-019-0188-4},
}

@inproceedings{4,
  title={An introductory survey on attention mechanisms in NLP problems},
  author={Hu, D.},
  booktitle={Intelligent Systems and Applications: Proceedings of the 2019 Intelligent Systems Conference (IntelliSys) Volume 2},
  pages={432--448},
  year={2020},
  organization={Springer International Publishing},
}

@article{6,
  title={A Security Model Based on LightGBM and Transformer to Protect Healthcare Systems From Cyberattacks},
  author={Ghourabi, A.},
  journal={IEEE Access},
  volume={10},
  pages={48890--48903},
  year={2022},
  doi={10.1109/ACCESS.2022.3172432},
}

@article{7,
  title={Fake news detection based on a hybrid BERT and LightGBM models},
  author={Essa, E. and Omar, K. and Alqahtani, A.},
  journal={Complex \& Intelligent Systems},
  pages={1--12},
  year={2023},
}

@article{8,
  title={Detection and Classification of mental illnesses on social media using RoBERTa},
  author={Murarka, A. and Radhakrishnan, B. and Ravichandran, S.},
  journal={arXiv preprint arXiv:2011.11226},
  year={2020},
}

@article{5,
  title={COVID-Twitter-BERT: A Natural Language Processing Model to Analyse COVID-19 Content on Twitter},
  journal={arXiv preprint arXiv:2005.07503},
  year={2020},
}

@inproceedings{10,
  title={Optuna: A next-generation hyperparameter optimization framework},
  author={Akiba, T. and Sano, S. and Yanase, T. and Ohta, T. and Koyama, M.},
  booktitle={Proceedings of the 25th ACM SIGKDD international conference on knowledge discovery \& data mining},
  pages={2623--2631},
  year={2019},
}

@article{9,
  author    = {Yinhan Liu and Myle Ott and Naman Goyal and Jingfei Du and Mandar Joshi and Danqi Chen and Omer Levy and Mike Lewis and Luke Zettlemoyer and Veselin Stoyanov},
  title     = {RoBERTa: {A} Robustly Optimized {BERT} Pretraining Approach},
  journal   = {arXiv preprint arXiv:1907.11692},
  year      = {2019},
  url       = {http://arxiv.org/abs/1907.11692},
  archivePrefix = {arXiv},
  eprint    = {1907.11692},
  bibsource = {dblp computer science bibliography, https://dblp.org}
}
\end{document}